\documentclass[runningheads]{llncs}
\usepackage[hidelinks]{hyperref}
\usepackage[T1]{fontenc}
\usepackage{booktabs}
\usepackage{multirow}
\usepackage{graphicx}
\usepackage{orcidlink}
\usepackage{cite}
\usepackage{enumitem}
\setlist{nosep}
\makeatletter
\newcommand{\printfnsymbol}[1]{%
  \textsuperscript{\@fnsymbol{#1}}%
}
\makeatother
\begin{document}
\title{Exploring Multi-Modal Large Language Models and Two-Stage Fine-Tuning for Fashion Image Retrieval}
\titlerunning{Exploring MLLMs and Two-Stage Fine-Tuning for Fashion Image Retrieval}
%
\author{Nguyen Cao Hoang\inst{1, 2}\orcidlink{0009-0003-0459-730X}\thanks{These authors contributed equally to this work.} \and
Hoang Bui Le\inst{1,2}\orcidlink{0009-0003-1162-234X}\printfnsymbol{1} \and
Nam Vo Hoang\inst{1,2}\orcidlink{0009-0000-2835-3711}\printfnsymbol{1} \and Trung-Nghia Le\inst{1, 2}\orcidlink{0000-0002-7363-2610}\thanks{Corresponding author.}}
\authorrunning{Nguyen Cao Hoang et al.}
%
\institute{University of Science, VNU-HCM, Ho Chi Minh, Vietnam \and
  Vietnam National University, Ho Chi Minh, Vietnam\\
\email{\{23125064, 23125057, 23125040\}@student.hcmus.edu.vn}
\email{ltnghia@fit.hcmus.edu.vn}
}
\maketitle              

\begin{abstract}
    Composed image retrieval retrieves a target image using a composed query of a reference image and a modified text description. In the fashion domain, this task requires understanding subtle attribute variations such as color, pattern, and texture. However, existing approaches face limitations due to scarce annotated data and simplistic negative sampling. We propose a novel framework that integrates a multi-modal large language model (LLaVA) to generate attribute-aware triplets and introduces a two-stage fine-tuning strategy to enhance contrastive learning. We leverage pretrained vision-language models, such as CLIP-ViT/B32, to generate and concatenate sentence-level prompts with the relative caption and to scale the number of negatives using static representations. Experimental results demonstrate enhanced compositional reasoning and improved fine-grained retrieval behavior, underscoring the feasibility and potential of the proposed framework for fashion retrieval.

\end{abstract}

\section{Introduction}

Composed image retrieval (CIR) is a challenging retrieval task where a query consists of a reference image and a relative caption, aiming to locate a target image that reflects the described modifications while retaining visual similarity to the reference \cite{liu2021, baldrati2022}. Within this paradigm, Fashion image retrieval (FIR) emerges as a specialized and fine-grained instance, tailored for fashion applications such as e-commerce, personalized shopping, and virtual try-on \cite{pal2023, liu2021}. Unlike general CIR, FIR demands precise interpretation of subtle visual attributes, such as color tone, texture, pattern, and fit, based on detailed, multi-attribute user queries (e.g., “make this dress pastel blue with long sleeves and a floral pattern”).

Despite recent advances, FIR remains limited by shallow visual understanding and inefficient contrastive learning. Models such as CLIP~\cite{alec2021}, while powerful for general vision-language alignment, primarily capture global semantics and often overlook subtle, fine-grained attributes crucial in fashion, for instance, intricate lace textures or nuanced silhouette variations. Furthermore, the scarcity and ambiguity of annotated training data prevent contrastive objectives from effectively learning detailed visual distinctions. As a result: (i) models lack sufficient diverse positive examples, and (ii) the common practice of using random in-batch negatives fails to expose the model to genuinely hard, visually similar candidates.

To overcome these challenges, we present a framework that enhances visual representation learning through enriched image captioning and improved negative sampling. Specifically, we employ the multimodal LLM (LLaVA)~\cite{liu2023} to generate high-quality, attribute-aware captions that enrich visual-textual alignment and mitigate data sparsity. Additionally, we introduce a two-stage fine-tuning strategy incorporating both coarse and hard-negative alignment to strengthen discriminative learning. Together, these components enable more robust, fine-grained feature representations for fashion retrieval. Experimental results on the FashionIQ dataset \cite{wu2020} demonstrate enhanced compositional reasoning and improved fine-grained retrieval behavior, underscoring the feasibility and potential of the proposed framework for fashion retrieval.

Our contributions are as follows:
\begin{itemize}
    \item We employ LLaVA to generate high-quality, attribute-aware captions and triplets, enriching image-text alignment and mitigating the shortage of annotated examples.
    \item We design a two-stage fine-tuning strategy that combines coarse alignment with hard-negative sampling to strengthen discriminative learning and improve fine-grained retrieval accuracy.
    \item We integrate sentence-level prompting with relative captions using pretrained vision-language models (e.g., BLIP-2), enhancing compositional reasoning and interpretability in composed queries.
\end{itemize}

\section{Related Work}

\textbf{Composed image retrieval (CIR)} combines a query image with modifying text to retrieve target images, necessitating effective fusion of visual and textual modalities. Recent works leverage large-scale vision-language models: Liu et al.~\cite{liu2021} introduced CIRPLANT using transformer-based adaptation, while Baldrati et al.~\cite{baldrati2022,baldrati2023} exploited CLIP for robust feature fusion. Xu et al.~\cite{xu2023} proposed ComqueryFormer, a unified transformer architecture with global-local alignment, and Zhao et al.~\cite{zhao2022} incorporated progressive learning with adaptive weighting for hybrid queries. Additional innovations include sentence-level prompting~\cite{bai2023}, zero-shot methods~\cite{saito2023}, and context-aware mapping techniques~\cite{tang2024}, as well as extensions to video retrieval~\cite{ventura2024}. 

The challenge of bridging the semantic gap between low-level visual features and high-level fashion concepts has been addressed in \textbf{fashion image retrieval (FIR)}. Research highlights low-level features and optimisation algorithms for semantic recognition~\cite{hare2006,karmokar2013}, while later advancements include semantic fusion networks~\cite{liu2020} and compositional approaches~\cite{valle2018} to capture outfit constituents. Interactive retrieval systems supported by datasets such as Fashion IQ~\cite{wu2020} have emerged, utilizing methods like mix attention-based CNNs for brand logo recognition~\cite{liu2021} and memory-based models for multi-turn feedback~\cite{pal2023}. These studies underscore the importance of semantic understanding and iterative user feedback in refining retrieval results and enhancing overall retrieval efficiency.

\textbf{Negative sampling} plays a vital role in contrastive learning for image retrieval. Feng et al.~\cite{feng2024} used multi-modal language models to generate triplets for CIR, addressing positive data scarcity. Zhou and Li~\cite{zhou2024} proposed a coarse-to-fine alignment framework for cross-modal image retrieval, improving performance through targeted sampling. Contrastive learning approaches, such as SimCLR~\cite{chen2020} and non-parametric instance discrimination~\cite{wu2018}, have shown that augmentations and effective sampling of negatives are essential for learning robust representations. Additionally, conditional negative sampling~\cite{wu2020} enhances feature transferability to new distributions, while contrastive hashing with vision transformer~\cite{ren2022} improves retrieval performance by integrating hard negative samples.


\section{Methodology}

\subsection{Overview}
\begin{figure}[t!]
    \centering
    \includegraphics[width=\columnwidth]{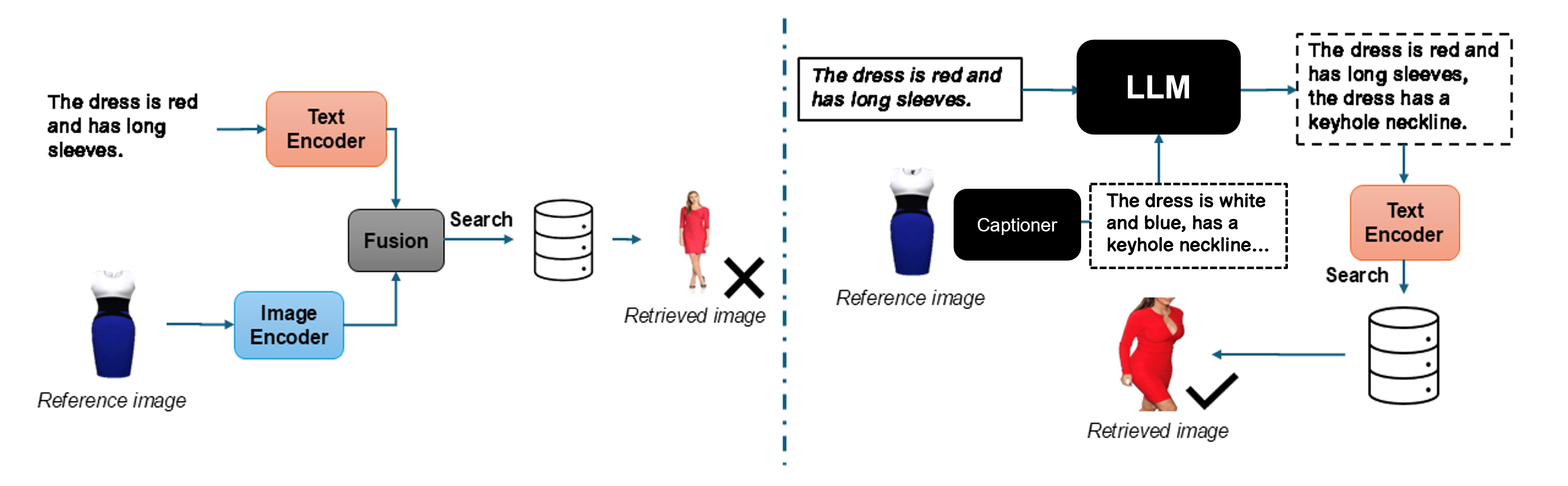}
    \caption{Our proposed framework (Right), compared with the standard CIR \cite{baldrati2023} (Left).}
    \label{fig:Pipeline}
    \vspace{-5mm}
\end{figure}
Our method builds upon the standard CIR framework \cite{baldrati2023} (Fig. \ref{fig:Pipeline}), where a query combines a reference image and a modification text to retrieve a visually similar target image reflecting the described changes. Like prior work, we employ a dual-encoder contrastive model: the query encoder jointly embeds the reference image and modification text, while the target encoder processes candidate images. Training optimizes cosine similarity between matching query–target pairs using an in-batch contrastive loss.

The key innovation is enhancing the reference image representation through LLaVA-generated, attribute-aware captions. These captions capture fine-grained visual details, such as color, pattern, texture, and style, often missed by CLIP’s global features. The generated caption is concatenated with the modification text to form a richer, contextually grounded textual input, improving fusion and alignment for fine-grained retrieval.

\subsection{Enhanced Caption Generation via LLaVA}
We enrich reference image descriptions using LLaVA, a vision-language model that integrates a vision encoder and language model end-to-end ~\cite{liu2023}. To generate detailed, context-aware captions highlighting fine-grained fashion attributes, we adopt a two-step prompting strategy:

\subsubsection{Image-Conditioned Prompting}
For each reference image \(r\) and target image \(t\), we input them into LLaVA alongside a structured prompt: \textit{``Describe this fashion item in detail, focusing on color, pattern, texture, and style. Highlight any distinctive elements.''} This yields caption \(C_r\) and \(C_t\), which captures fine-grained visual attributes (e.g., ``a knee-length dress with floral embroidery on a navy blue silk base'').

\subsubsection{Modified Text Synthesis}
To synthesize the modified text \(t\), we concatenate \(C_r\) with the relative caption provided in the dataset (e.g., ``make it pastel blue'') and feed this into LLaVA with a follow-up prompt: \textit{``Generate a concise instruction that modifies the original description based on the given change.''} This produces a context-aware modified caption (e.g., ``Change the navy blue silk dress to a pastel blue tone while retaining the floral embroidery''). This process ensures that \(t\) explicitly references attributes in \(C_r\), reducing ambiguity.

\subsection{Two-Stage Fine-Tuning with Augmented Triplets}
We adopt a two-stage training framework to leverage both human-annotated and synthetic triplets:
\begin{figure}[thbp]
    \centering
    \includegraphics[width=\columnwidth]{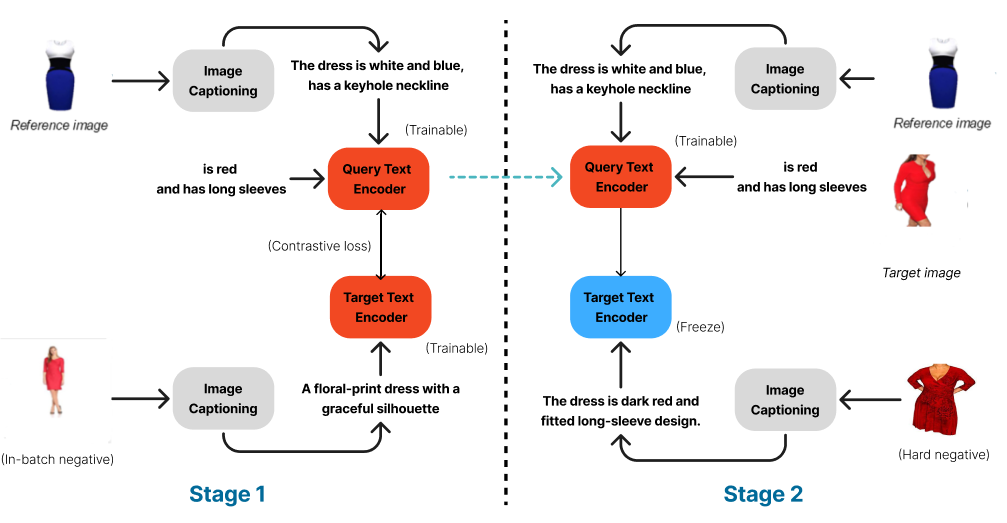}
    \caption{Overview of our two-stage fashion retrieval training Pipeline.}
    \label{fig:Stage 1}
\end{figure}
\begin{itemize}
    \item \textbf{Stage 1 (Coarse Alignment):} The query and target encoders are jointly trained with in-batch negatives to align reference images and modification texts. The query encoder combines CLIP-ViT features and LLaVA captions with relative text embeddings, enabling attribute-aware feature learning.
    \item \textbf{Stage 2 (Refined Alignment):} The target encoder is frozen while the query encoder is fine-tuned using hard negatives—samples with high similarity but mismatched attributes (e.g., same category, different color). This enhances the model’s ability to capture subtle attribute differences for fine-grained fashion retrieval.
\end{itemize}

\subsection{Negative Sampling Strategy}
To address the scarcity of negative samples, we adopt a \textbf{hybrid negative sampling strategy} inspired by Feng \textit{et al.}~\cite{feng2024}. Each training triplet is represented as $(c_t, c_r, t_u)$, where $c_t$ denotes the \textit{target caption}, $c_r$ the \textit{reference caption} generated by \textbf{LLaVA}, and $t_u$ the \textit{user modification text} describing the desired change. We define three complementary negative types:

\begin{itemize}
    \item \textbf{In-Batch Negatives:} Drawn from other samples within the same mini-batch, providing baseline diversity by contrasting each query $(c_r, t_u)$ against all non-matching $c_t$.
    
    \item \textbf{Synthetic Negatives:} Plausible but incorrect targets generated by perturbing the target caption $c_t$. For example, attribute terms within $c_t$ are modified (e.g., replacing ``pastel blue'' with ``emerald green''), producing semantically close but mismatched descriptions. This challenges the model’s ability to discriminate fine-grained differences.
    
    \item \textbf{Augmented Negatives:} Formed by cross-combining elements from different triplets, such as pairing an unrelated $c'_t$ with a similar $c_r$ or $t_u$, generating hard and diverse negatives that reduce spurious correlations.
\end{itemize}

This hybrid strategy exposes the model to both easy and hard negatives, thereby strengthening contrastive alignment and improving sensitivity to subtle attribute variations.

\section{Experiments}
\subsection{Implementation Details}

We used CLIP-ViT-B/32 for image encoding and LLaVA-v1.5-13b-3GB for text encoding. The fused image-text representations are processed through a 2-layer transformer to produce query embeddings. We froze CLIP text encoder, used to encode target images captioning from LLaVa consistently across all experiments.

\begin{table}[t!]
\centering
\caption{Performance of our proposed method compared with SOTA methods. The best and second best methods are shown in \textbf{bold} and \textit{italic}.}
\label{tab:comparison}
\begin{tabular}{lcccccccc}
\toprule
\multirow{2}{*}{\textbf{Method}} &
\multicolumn{2}{c}{\textbf{Dress}} &
\multicolumn{2}{c}{\textbf{Shirt}} &
\multicolumn{2}{c}{\textbf{Top-Tee}} &
\multicolumn{2}{c}{\textbf{Average}} \\
\cmidrule(lr){2-3} \cmidrule(lr){4-5} \cmidrule(lr){6-7} \cmidrule(lr){8-9}
 & R@10 & R@50 & R@10 & R@50 & R@10 & R@50 & R@10 & R@50 \\
\midrule
CompoDiff~\cite{gu2024} & 40.65 & 57.14 & 36.87 & 57.39 & 43.93 & 61.17 & 40.48 & 58.57 \\
SPRC~\cite{Bai2023SentenceLevelPrompts} & 47.80 & \textit{72.70} & 55.84 & 74.37 & 58.89 & 78.99 & 54.17 & 75.35 \\
MAPNet~\cite{Shi2025MultiSchemaProximity} & \textbf{51.17} & \textbf{74.12} & \textit{56.37} & \textit{75.17} & \textbf{59.56} & \textit{79.30} & \textbf{55.70} & \textbf{76.20} \\
\textbf{Ours} & {\textit{49.62}} & {71.83} & \textbf{57.02} & \textbf{75.66} & {\textit{59.31}} & \textbf{79.50} & {\textit{55.32}} & {\textit{75.67}} \\
\bottomrule
\end{tabular}

\end{table}

\subsection{Experimental Settings}
All experiments were performed on the \textit{FashionIQ} ~\cite{wu2020}, a natural language-based interactive fashion product retrieval dataset. It contains 77,684 images crawled from Amazon.com, covering three categories: Dresses, Tops \& Tees, and Shirts. Among the 46,609 training images, there are 18,000 image pairs. Each pair is accompanied by an average of two natural language sentences that describe one or multiple visual properties to modify in the reference image, such as ``is shiny'' or ``is blue in color and floral, and with white base.''

We employed \textbf{\(\mathit{Recall@K (R@K)}\)}~\cite{patel2022} as the primary evaluation metric, which measures the proportion of queries for which the retrieved top \(K\) images include the correct target image. 


\subsection{Experimental Results}
\subsubsection{Comparison with SOTA Methods}
On the full FashionIQ dataset, our CLIP-4CIR model with LLaVA-enhanced captioning achieves better quantitative results at Table~\ref{tab:comparison} than coarse-grained baselines like standard CIR models (e.g., CompoDiff~\cite{gu2024} and SPRC~\cite{Bai2023SentenceLevelPrompts}), though it remains below recent SOTA method (e.g., MAPNet~\cite{Shi2025MultiSchemaProximity}). The framework demonstrates stronger attribute expressiveness and compositional reasoning, particularly for fine-grained, multi-attribute queries. While minor alignment noise arises from the linguistic variability of generated captions, these enriched descriptions enhance visual-text alignment and retrieval interpretability. Overall, the results confirm the effectiveness of integrating multimodal caption refinement to advance fine-grained fashion retrieval.

\subsubsection{Qualitative Results}

\begin{figure}[t!]
    \centering
    \includegraphics[width=\textwidth]{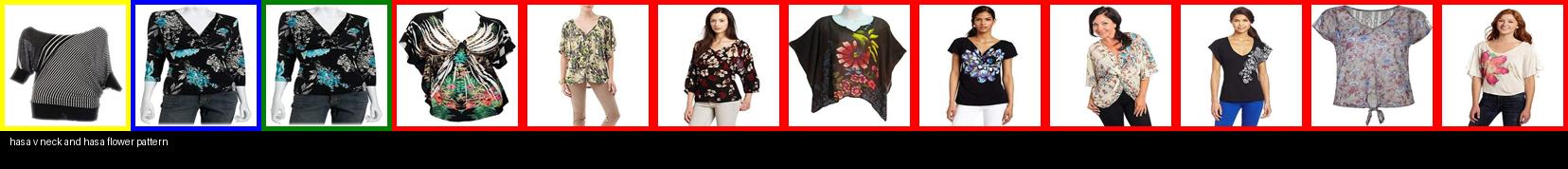}
    \includegraphics[width=\textwidth]{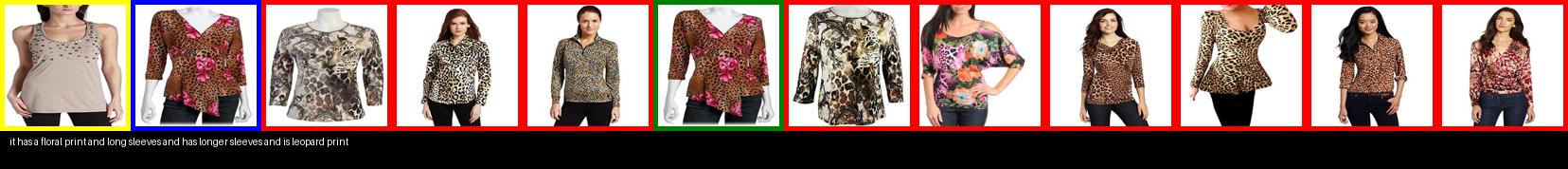}
    \includegraphics[width=\textwidth]{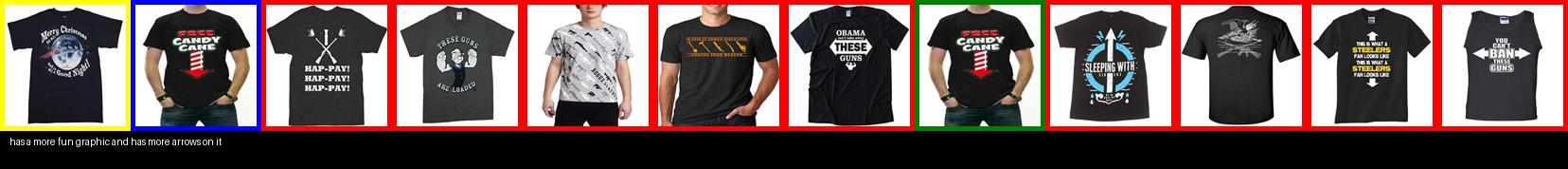}
    \includegraphics[width=\textwidth]{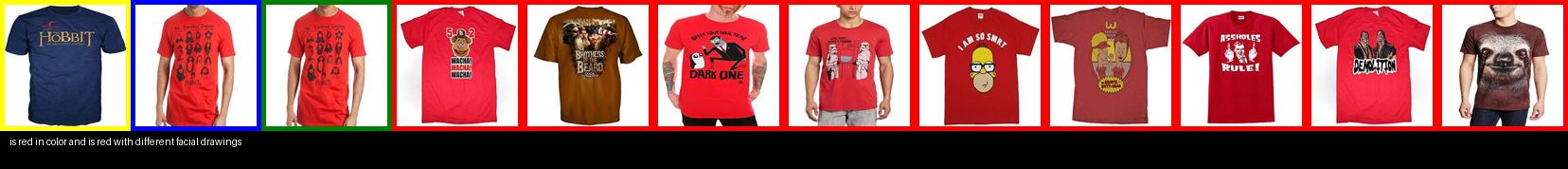}
    \caption{Illustrative positive examples of our method's performance. The captions for the queries are as follows: 
    (a) ``has a v neck and has a flower pattern'', 
    (b) ``it has a floral print and long sleeves and has longer sleeves and is leopard print'', 
    (c) ``has a more fun graphic and has more arrows on it'', and 
    (d) ``is red in color and is red with different facial drawings''.}
    \label{fig:success_examples}
\end{figure}

As shown in Figure~\ref{fig:success_examples}, our method accurately retrieves targets with clear geometry and distinct patterns, showing strong reasoning over attributes like neckline, sleeves, and graphics. LLaVA-enhanced captions further improve attention to fine-grained details beyond category-level semantics.

\begin{figure}[t!]
    \centering
    \includegraphics[width=\textwidth]{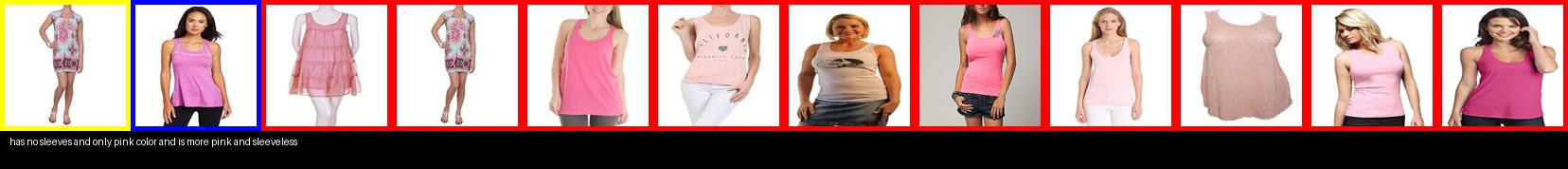}
    \includegraphics[width=\textwidth]{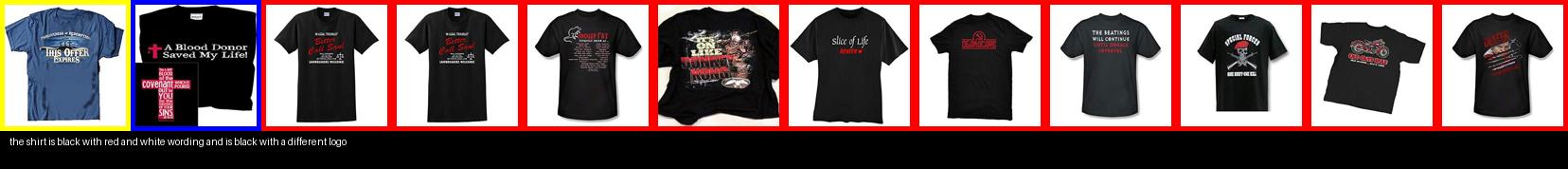}
    \includegraphics[width=\textwidth]{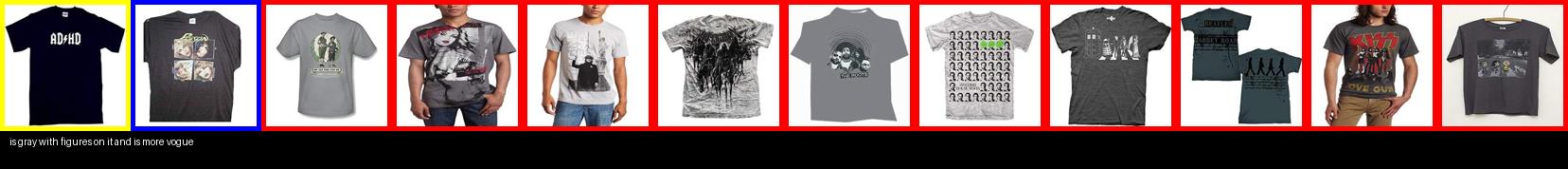}
    \includegraphics[width=\textwidth]{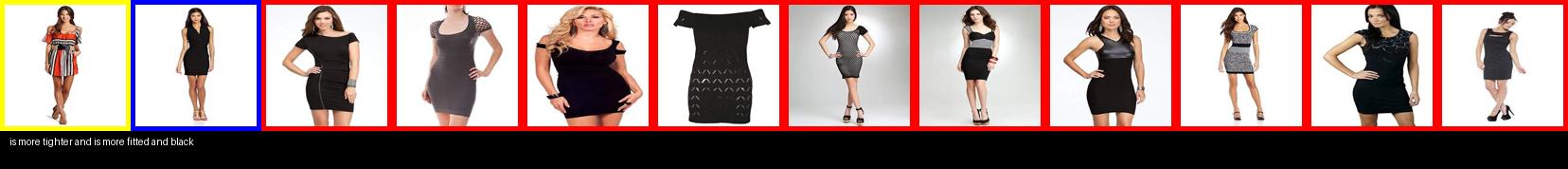}
    \caption{Illustrative negative examples of our method's performance. The captions for the queries are as follows:
    (a) ``has no sleeves and only pink color and is more pink and sleeveless'', 
    (b) ``it has a floral print and long sleeves and has longer sleeves and is leopard print'', 
    (c) ``is gray with figures on it and is more vogue'', and 
    (d) ``is more tighter and is more fitted and black''.}
    \label{fig:failure_examples}
\end{figure}

Conversely, Figure~\ref{fig:failure_examples} presents failure cases where the model struggles to resolve subtle variations, particularly for plain-colored garments or logo-based differences requiring pixel-level precision. These limitations suggest the need for spatial grounding or attention mechanisms to better localize fine-grained attributes.

This analysis highlights the model's strength in processing \textbf{composite attributes with distinctive visual signatures}, while revealing limitations in handling \textbf{generic color/shape descriptions} that dominate fashion datasets. The absence of spatial grounding mechanisms again exacerbates challenges in distinguishing near-identical plain-colored items.

\section{Discussion and Future Work}

\subsection{Discussion}

Although the proposed framework demonstrates encouraging qualitative behaviour and improved handling of complex, fine-grained queries, the overall quantitative performance still falls short of fully realizing the potential of fine-grained fashion retrieval. We attribute this observation to several interrelated factors:

\begin{itemize}
    \item \textbf{Caption Complexity and Alignment Noise.} LLaVA-generated captions, while rich in attribute-level detail, often exhibit higher linguistic variability and verbosity compared to human-annotated descriptions. This increases the semantic gap between textual and visual embeddings.
    
    \item \textbf{Absence of Spatial Grounding.} The model encodes images globally without explicit mechanisms to focus on attribute-relevant regions (e.g., sleeves, neckline, logos). As a result, retrieval performance degrades in cases where subtle, localized differences determine correctness.
    
    \item \textbf{Limited Fine-Tuning Scale.} Current results are derived from a partial dataset due to time and resource constraints. Training on limited samples restricts the diversity of negative pairs and constrains the model’s exposure to visual variations, limiting its generalization across broader retrieval scenarios.
    
    \item \textbf{Imbalance in Attribute Distribution.} Fashion datasets are skewed toward dominant features such as colour or pattern, while fine-grained attributes like fabric texture or subtle styling cues remain underrepresented. This imbalance can bias the model toward coarser visual signals, limiting precision in detailed queries.
\end{itemize}

Despite these limitations, the qualitative improvements observed in compositional reasoning and attribute-level retrieval underscore the potential of integrating large multimodal language models into composed image retrieval. These findings suggest that, with appropriate alignment strategies, the proposed framework can leverage the power of image captioning to bridge the gap between visual and linguistic semantics.

\subsection{Potential and Future Directions}

The findings reveal several promising directions for further development:

\begin{itemize}
    \item \textbf{Caption Refinement and Filtering.} Future work will explore structured prompting and linguistic post-processing (e.g., attribute extraction and simplification) to mitigate noise introduced by free-form LLaVA captions. A filtering pipeline that retains only dataset-aligned attribute terms may enhance embedding consistency.
    \item \textbf{Spatial Grounding and Attention.} Integrating spatial attention mechanisms or region-level feature pooling could enable the model to localize modifications more accurately. Such grounding is critical for resolving subtle distinctions, such as sleeve types or small logos, that current global embeddings may overlook.
\end{itemize}

Overall, while the current quantitative metrics remain close to baseline, the observed qualitative gains in compositional reasoning indicate significant potential. With continued optimization in caption alignment, grounding, and sampling strategies, the proposed framework has the capacity to advance fine-grained fashion retrieval beyond existing contrastive learning paradigms.

\section{Conclusion}
This study demonstrates the feasibility of enhancing composed image retrieval through refined caption generation and contrastive learning. By integrating a large multimodal language model (LLaVA) into the retrieval pipeline, we introduce a novel mechanism for generating attribute-aware captions that capture fine-grained visual details often overlooked by traditional annotation methods. Preliminary experiments indicate that, while overall recall remains slightly below baseline levels, the proposed framework exhibits stronger performance in complex, multi-attribute queries, highlighting its potential for improved compositional reasoning and interpretability. The observed trade-off between detailed caption expressiveness and alignment stability underscores the importance of future research into structured prompting, linguistic filtering, and attribute-level supervision. It will serve a promising work to bridge the gap between synthetic caption-based training and real-world search behaviour.

\begin{credits}
\subsubsection{\ackname} 
This research is supported by research funding from Faculty of Information Technology, University of Science, Vietnam National University - Ho Chi Minh City.
\end{credits}

\bibliographystyle{splncs04}
\bibliography{references}

@misc{baldrati2023,
  author        = {Alberto Baldrati and Marco Bertini and Tiberio Uricchio and Alberto del Bimbo},
  title         = {Composed Image Retrieval using Contrastive Learning and Task-oriented CLIP-based Features},
  year          = {2023},
  url           = {https://arxiv.org/abs/2308.11485},
  eprint        = {2308.11485},
  archiveprefix = {arXiv},
  primaryclass  = {cs.CV},
}

@misc{alec2021,
  author       = {Alec Radford and Jong Wook Kim and Chris Hallacy and Aditya Ramesh and Gabriel Goh and Sandhini Agarwal and Girish Sastry and Amanda Askell and Pamela Mishkin and Jack Clark and Gretchen Krueger and Ilya Sutskever},
  title        = {Learning Transferable Visual Models From Natural Language Supervision},
  year         = {2021},
  archivePrefix= {arXiv},
  eprint       = {2103.00020},
  primaryClass = {cs.CV},
  url          = {https://arxiv.org/abs/2103.00020},
  note         = {arXiv preprint arXiv:2103.00020}
}

@inproceedings{baldrati2022,
  author    = {Baldrati, Alberto and Bertini, Marco and Uricchio, Tiberio and Del Bimbo, Alberto},
  title     = {Conditioned and Composed Image Retrieval Combining and Partially Fine-Tuning CLIP-Based Features},
  year      = {2022},
  pages     = {4959--4968},
  month     = jun,
  booktitle = {Proceedings of the IEEE/CVF Conference on Computer Vision and Pattern Recognition (CVPR) Workshops},
}

@misc{gu2024,
  author        = {Geonmo Gu and Sanghyuk Chun and Wonjae Kim and HeeJae Jun and Yoohoon Kang and Sangdoo Yun},
  title         = {CompoDiff: Versatile Composed Image Retrieval With Latent Diffusion},
  year          = {2024},
  url           = {https://arxiv.org/abs/2303.11916},
  eprint        = {2303.11916},
  archiveprefix = {arXiv},
  primaryclass  = {cs.CV},
}

@inproceedings{hare2006,
  author       = {Hare, Jonathon S and Lewis, Paul H and Enser, Peter GB and Sandom, Christine J},
  title        = {Mind the gap: Another look at the problem of the semantic gap in image retrieval},
  year         = {2006},
  volume       = {6073},
  pages        = {75--86},
  booktitle    = {Multimedia Content Analysis, Management, and Retrieval 2006},
  organization = {SPIE},
}

@misc{wu2020,
  author        = {Hui Wu and Yupeng Gao and Xiaoxiao Guo and Ziad Al-Halah and Steven Rennie and Kristen Grauman and Rogerio Feris},
  title         = {Fashion IQ: A New Dataset Towards Retrieving Images by Natural Language Feedback},
  year          = {2020},
  url           = {https://arxiv.org/abs/1905.12794},
  eprint        = {1905.12794},
  archiveprefix = {arXiv},
  primaryclass  = {cs.CV},
}

@article{karmokar2013,
  author    = {Karmokar, Punam R and Parekh, Ranjan},
  title     = {Recognition of Semantic Content in Image and Video},
  journal   = {International Journal of Computer Applications},
  year      = {2013},
  volume    = {73},
  number    = {15},
  publisher = {Citeseer},
}

@misc{zhou2024,
  author        = {Lifeng Zhou and Yuke Li},
  title         = {Coarse-to-fine Alignment Makes Better Speech-image Retrieval},
  year          = {2024},
  url           = {https://arxiv.org/abs/2408.13119},
  eprint        = {2408.13119},
  archiveprefix = {arXiv},
  primaryclass  = {cs.CL},
}

@article{liu2020,
  author    = {Liu, An-An and Zhang, Ting and Song, Dan and Li, Wenhui and Zhou, Ming},
  title     = {FRSFN: A semantic fusion network for practical fashion retrieval},
  journal   = {Multimedia Tools and Applications},
  year      = {2021},
  volume    = {80},
  pages     = {17169--17181},
  publisher = {Springer},
}

@article{liu2023,
  author  = {Liu, Haotian and Li, Chunyuan and Wu, Qingyang and Lee, Yong Jae},
  title   = {Visual instruction tuning},
  journal = {Advances in neural information processing systems},
  year    = {2023},
  volume  = {36},
  pages   = {34892--34916},
}

@inproceedings{liu2021,
  author    = {Liu, Zheyuan and Rodriguez-Opazo, Cristian and Teney, Damien and Gould, Stephen},
  title     = {Image Retrieval on Real-Life Images With Pre-Trained Vision-and-Language Models},
  year      = {2021},
  pages     = {2125-2134},
  month     = oct,
  booktitle = {Proceedings of the IEEE/CVF International Conference on Computer Vision (ICCV)},
}

@inproceedings{pal2023,
  author    = {Pal, Anwesan and Wadhwa, Sahil and Jaiswal, Ayush and Zhang, Xu and Wu, Yue and Chada, Rakesh and Natarajan, Pradeep and Christensen, Henrik I.},
  title     = {FashionNTM: Multi-turn Fashion Image Retrieval via Cascaded Memory},
  year      = {2023},
  pages     = {11323-11334},
  month     = {October},
  booktitle = {Proceedings of the IEEE/CVF International Conference on Computer Vision (ICCV)},
}

@article{ren2022,
  author   = {Ren, Xiuxiu and Zheng, Xiangwei and Zhou, Huiyu and Liu, Weilong and Dong, Xiao},
  title    = {Contrastive hashing with vision transformer for image retrieval},
  journal  = {International Journal of Intelligent Systems},
  year     = {2022},
  volume   = {37},
  number   = {12},
  pages    = {12192-12211},
  doi      = {https://doi.org/10.1002/int.23082},
  url      = {https://onlinelibrary.wiley.com/doi/abs/10.1002/int.23082},
  eprint   = {https://onlinelibrary.wiley.com/doi/pdf/10.1002/int.23082},
}

@inproceedings{saito2023,
  author    = {Saito, Kuniaki and Sohn, Kihyuk and Zhang, Xiang and Li, Chun-Liang and Lee, Chen-Yu and Saenko, Kate and Pfister, Tomas},
  title     = {Pic2Word: Mapping Pictures to Words for Zero-Shot Composed Image Retrieval},
  year      = {2023},
  pages     = {19305-19314},
  month     = jun,
  booktitle = {Proceedings of the IEEE/CVF Conference on Computer Vision and Pattern Recognition (CVPR)},
}

@article{tang2024,
  author       = {Tang, Yuanmin and Yu, Jing and Gai, Keke and Zhuang, Jiamin and Xiong, Gang and Hu, Yue and Wu, Qi},
  title        = {Context-I2W: Mapping Images to Context-Dependent Words for Accurate Zero-Shot Composed Image Retrieval},
  journal      = {Proceedings of the AAAI Conference on Artificial Intelligence},
  year         = {2024},
  volume       = {38},
  number       = {6},
  pages        = {5180-5188},
  month        = mar,
  doi          = {10.1609/aaai.v38i6.28324},
  url          = {https://ojs.aaai.org/index.php/AAAI/article/view/28324},
  abstractnote = {Different from the Composed Image Retrieval task that requires expensive labels for training task-specific models, Zero-Shot Composed Image Retrieval (ZS-CIR) involves diverse tasks with a broad range of visual content manipulation intent that could be related to domain, scene, object, and attribute. The key challenge for ZS-CIR tasks is to learn a more accurate image representation that has adaptive attention to the reference image for various manipulation descriptions. In this paper, we propose a novel context-dependent mapping network, named Context-I2W, for adaptively converting description-relevant Image information into a pseudo-word token composed of the description for accurate ZS-CIR. Specifically, an Intent View Selector first dynamically learns a rotation rule to map the identical image to a task-specific manipulation view. Then a Visual Target Extractor further captures local information covering the main targets in ZS-CIR tasks under the guidance of multiple learnable queries. The two complementary modules work together to map an image to a context-dependent pseudo-word token without extra supervision. Our model shows strong generalization ability on four ZS-CIR tasks, including domain conversion, object composition, object manipulation, and attribute manipulation. It obtains consistent and significant performance boosts ranging from 1.88% to 3.60% over the best methods and achieves new state-of-the-art results on ZS-CIR. Our code is available at https://anonymous.4open.science/r/Context-I2W-4224/.},
}

@misc{chen2020,
  author        = {Ting Chen and Simon Kornblith and Mohammad Norouzi and Geoffrey Hinton},
  title         = {A Simple Framework for Contrastive Learning of Visual Representations},
  year          = {2020},
  url           = {https://arxiv.org/abs/2002.05709},
  eprint        = {2002.05709},
  archiveprefix = {arXiv},
  primaryclass  = {cs.LG},
}

@inproceedings{valle2018,
  author    = {Valle, Dan and Ziviani, Nivio and Veloso, Adriano},
  title     = {Effective Fashion Retrieval Based on Semantic Compositional Networks},
  year      = {2018},
  volume    = {},
  number    = {},
  pages     = {1-8},
  doi       = {10.1109/IJCNN.2018.8489494},
  booktitle = {2018 International Joint Conference on Neural Networks (IJCNN)},
}

@article{ventura2024,
  author    = {Ventura, Lucas and Yang, Antoine and Schmid, Cordelia and Varol, Gül},
  title     = {CoVR-2: Automatic Data Construction for Composed Video Retrieval},
  journal   = {IEEE Transactions on Pattern Analysis and Machine Intelligence},
  year      = {2024},
  volume    = {46},
  number    = {12},
  pages     = {11409--11421},
  month     = dec,
  doi       = {10.1109/tpami.2024.3463799},
  url       = {http://dx.doi.org/10.1109/TPAMI.2024.3463799},
  issn      = {1939-3539},
  publisher = {Institute of Electrical and Electronics Engineers (IEEE)},
}

@article{xu2023,
  author     = {Xu, Yahui and Bin, Yi and Wei, Jiwei and Yang, Yang and Wang, Guoqing and Shen, Heng Tao},
  title      = {Multi-Modal Transformer With Global-Local Alignment for Composed Query Image Retrieval},
  journal    = {Trans. Multi.},
  year       = {2023},
  volume     = {25},
  number     = {1},
  pages      = {8346--8357},
  month      = jan,
  doi        = {10.1109/TMM.2023.3235495},
  url        = {https://doi.org/10.1109/TMM.2023.3235495},
  issn       = {1520-9210},
  publisher  = {IEEE Press},
  issue_date = {2023},
  numpages   = {12},
}

@misc{bai2023,
  author        = {Yang Bai and Xinxing Xu and Yong Liu and Salman Khan and Fahad Khan and Wangmeng Zuo and Rick Siow Mong Goh and Chun-Mei Feng},
  title         = {Sentence-level Prompts Benefit Composed Image Retrieval},
  year          = {2023},
  url           = {https://arxiv.org/abs/2310.05473},
  eprint        = {2310.05473},
  archiveprefix = {arXiv},
  primaryclass  = {cs.CV},
}

@misc{patel2022,
  author        = {Yash Patel and Giorgos Tolias and Jiri Matas},
  title         = {Recall@k Surrogate Loss with Large Batches and Similarity Mixup},
  year          = {2022},
  url           = {https://arxiv.org/abs/2108.11179},
  eprint        = {2108.11179},
  archiveprefix = {arXiv},
  primaryclass  = {cs.CV},
}

@misc{zhao2022,
  author        = {Yida Zhao and Yuqing Song and Qin Jin},
  title         = {Progressive Learning for Image Retrieval with Hybrid-Modality Queries},
  year          = {2022},
  url           = {https://arxiv.org/abs/2204.11212},
  eprint        = {2204.11212},
  archiveprefix = {arXiv},
  primaryclass  = {cs.CV},
}

@misc{feng2024,
  author        = {Zhangchi Feng and Richong Zhang and Zhijie Nie},
  title         = {Improving Composed Image Retrieval via Contrastive Learning with Scaling Positives and Negatives},
  year          = {2024},
  url           = {https://arxiv.org/abs/2404.11317},
  eprint        = {2404.11317},
  archiveprefix = {arXiv},
  primaryclass  = {cs.CV},
}

@misc{wu2018,
  author        = {Zhirong Wu and Yuanjun Xiong and Stella Yu and Dahua Lin},
  title         = {Unsupervised Feature Learning via Non-Parametric Instance-level Discrimination},
  year          = {2018},
  url           = {https://arxiv.org/abs/1805.01978},
  eprint        = {1805.01978},
  archiveprefix = {arXiv},
  primaryclass  = {cs.CV},
}

@article{Bai2023SentenceLevelPrompts,
  author    = {Yang Bai and Xinxing Xu and Yong Liu and Salman Khan and Fahad Khan and Wangmeng Zuo and Rick Siow Mong Goh and Chun-Mei Feng},
  title     = {Sentence-level Prompts Benefit Composed Image Retrieval},
  journal   = {arXiv preprint},
  year      = {2023},
  archivePrefix = {arXiv},
  eprint    = {2310.05473},
  primaryClass = {cs.CV}
}

@inproceedings{Shi2025MultiSchemaProximity,
  author    = {Jiangming Shi and Xiangbo Yin and Yeyun Chen and Yachao Zhang and Zhizhong Zhang and Yuan Xie and Yanyun Qu},
  title     = {Multi-Schema Proximity Network for Composed Image Retrieval},
  booktitle = {Proceedings of the IEEE/CVF International Conference on Computer Vision (ICCV) 2025},
  year      = {2025}
}
\end{document}